\documentclass{article}
\usepackage{spconf,amsmath,graphicx}

\usepackage{cite}
\usepackage{amsmath,amssymb,amsfonts}
\usepackage{algorithmic}
\usepackage{graphicx}
\usepackage{textcomp}
\usepackage{xcolor}

\usepackage{multirow}
\usepackage{hyperref}
\usepackage{booktabs}
\usepackage{subcaption}
\usepackage{comment}
\newcommand{\bs}{\mathbf}

\usepackage{booktabs, multicol, multirow,array} %
\newcolumntype{?}{!{\vrule width 1pt}}
\newcolumntype{|}{!{\vrule width 0.4pt}}
\usepackage{arydshln}

\title{Facial Emotion Recognition using Deep Residual Networks in Real-World Environments}
\name{Panagiotis Tzirakis$^{1}$, D\'enes Boros$^2$, Elnar Hajiyev$^3$, Bj\"orn W.\ Schuller$^{1,4}$}
\address{$^1$ GLAM -- Group on Language, Audio \& Music, Imperial College London, UK \\
         $^2$ R\&D Department, RealEyes, Budabest, Hungary \\
         $^3$ R\&D Department, RealEyes, London, UK \\
         $^4$ Chair of Embedded Intelligence for Health Care and Wellbeing, University of Augsburg, Germany\\
{\small \tt panagiotis.tzirakis12@imperial.ac.uk}}

\begin{document}
%\ninept
%
\maketitle
\begin{abstract}
Automatic affect recognition using visual cues is an important task towards a complete interaction between humans and machines. Applications can be found in tutoring systems and human computer interaction. A critical step towards that direction is facial feature extraction. In this paper, we propose a facial feature extractor model trained on an in-the-wild and massively collected video dataset provided by the RealEyes company. The dataset consists of a million labelled frames and $2,616$ thousand subjects. As temporal information is important to the emotion recognition domain, we utilise LSTM cells to capture the temporal dynamics in the data. To show the favourable properties of our pre-trained model on modelling facial affect, we use the RECOLA database, and compare with the current state-of-the-art approach. Our model provides the best results in terms of concordance correlation coefficient.
\end{abstract}
\section{Introduction}
\label{sec:intro}

Automatic analysis of facial affect is a vital process in human-to-human interactions. Currently,  intelligent systems, such as robots and virtual humans, try to use emotion recognition models to make the interaction with humans more \textit{natural}. To this end, such systems should automatically sense and adapt their responses according to the human behaviour. One application can be found in an automatic tutoring system, where the system adjusts the level of the tutorial depending on the user's affective state, such as excitement or boredom. In another application, a fatigue state can be predicted by utilising affect states~\cite{ji2004real}.

% However, recognising human affect is a very challenging task. First, the duration of human emotions varies significantly and depends on a specific situation and the person. In addition, emotions are expressed differently among different individuals~\cite{anagnostopoulos2015features} and cultures. That is, for the same circumstances different persons can react in a very different manner. The difficulty of affect recognition is further enhanced in real-world environments where uncontrolled conditions are entailed, and subjects can perform any kind of expression.

For the past twenty years, a number of studies have been proposed in the literature to recognise emotions~\cite{tzirakis2017multimodal, kollias2018deep}. Most of them were focused on predicting the emotions in posed settings, namely, controlled environments. However, a shift has been observed in the affective community towards the prediction of the affective state of individuals in real-world environments~\cite{tzirakis21affect}. The reason for this shift is the observation that posed facial expressions can vary considerably in naturalistic behaviour~\cite{corneanu2016survey, sariyanidi2015automatic, tzirakis2019book}.

Deep Neural Networks (DNNs) have been widely used in the affective computing domain and have shown to provide state-of-the-art-results in several emotion recognition tasks~\cite{kollias2018deep, tzirakis21affect}. In particular, Convolution Neural Networks (CNNs) are widely used to extract features from images. As emotions are expressed in temporal boundaries, it is important for the models to capture the contextual dynamics in the signal. To this end, Recurrent Neural Networks (RNNs) are utilised that can effectively model temporal patterns in the data.

In this paper, we present a pre-trained deep residual network of 50 layers (ResNet-50)~\cite{he2016deep} that is trained in an \textit{end-to-end} manner utilising a real-world industrial dataset collected by the RealEyes company~\footnote{https://www.realeyesit.com/}. The dataset is comprised of more than a million images and more than $2,616$ thousand subjects for categorical emotion recognition, and in particular predicting eight emotions. To the best of our knowledge, we are the first to train a model on thousands of subjects for real-world automatic emotion recognition. The main drawback of this dataset is the highly unbalanced number of labels, which we tackle using boosting methods. 
We show the favourable properties of our model to recognise the affective state by utilising the REmote COLlaborative and  Affective (RECOLA) database~\cite{Ringeval13-ITR}. More particularly, we train the ResNet-50 in four different ways: (i) from scratch, i.\,e., random initialisation of the weights, (ii) utilising the pre-trained one from ImageNet~\cite{deng2009imagenet}, (iii) utilising the pre-trained one from AffWild database~\cite{kollias2018deep}, and (iv) using our pre-trained model. Our model surpasses the performance of the other three models in terms of Concordance Correlation Coefficient ($\rho_c$)~\footnote{Model is available with our toolkit \href{https://github.com/end2you/end2you}{End2You}}. 

% Finally, we observed a faster convergence of the optimisation of our model than the compared models.
\section{Related Work}
\label{sec:related_work}

Several studies have been proposed to model affect using visual cues. 
%For instance, Nicolaou et al.~\cite{nicolaou2012output} proposed a regression framework named Output-Associative Relevance Vector Machine (OA-RVM), which expands the RVM model so as to be able to learn non-linear dependencies from the input and the output. In a different study, Chen et al.~\cite{chen20153d} proposed to recognise dimensional emotion recognition by first restoring 3D facial models from 2D images. Then, they use a variant of the random-forest algorithm for the final prediction.
For instance, Huang et al.~\cite{huang2016deep} proposed a bi-stage learning framework for image-based emotion recognition. In particular, the authors combine DNNs and hypergraphs by first training the DNN for an emotion classification task, and then consider each node in the last fully connected layer as an attribute and use it to form a hyperedge in a hypergraph. In a more recent study, Kollias et al.~\cite{kollias2018deep} proposed an in-the-wild dataset, named Aff-Wild, that consists of $298$ videos of $200$ subjects, with a total length of more than $30$ hours. They trained end-to-end deep architectures consisting of CNNs, that spatially extract features from the image, and RNNs, so as the contextual dynamics in sequential data can be captured. The feature extraction models were based on the ResNet-50 and the VGG-Face/VGG-16~\cite{Parkhi15vggface} architectures, with two fully connected layers on top, before feeding the features to a 2-layer LSTM network. VGG produced the best results in their experiments. However, their models were trained on a dataset that consists of a small number of individuals (only 200), whereas our model was trained on more than thousand subjects. The interested reader is referenced to~\cite{ko2018brief} for more studies in the literature for visual-based emotion recognition.

\section{Methods: Model Design}

Our network comprises of two parts: (a) a feature extractor, that extracts spatial features from the images, and  (b) a contextual extractor that captures the temporal dynamics in the data. Our feature extractor is a ResNet-50. The extracted features are being fed to a 2-layer LSTM that captures the temporal information in the data. Finally, a 2-unit fully connected layer is inserted that represents the output (arousal, and valence) of the network. 
%Figure~\ref{fig:network} depicts our network.

% \begin{figure*}[h]
% \centering
% \includegraphics[width=16cm, height=4.5cm]{network}
% \caption{The network comprises of two parts: the feature extractor and the contextual extractor. Our feature extractor is a ResNet-50. The extracted features are used to feed a 2-layer LSTM layers that  captures the temporal dynamics in the data. (The input image shown is from the RECOLA database.)}
% \label{fig:network}
% \end{figure*}

% \subsection{Feature Extractor}
% \label{sec:resnet}
\textit{Feature Extractor}: In traditional face recognition an important step is feature extraction. This is performed by using hand-crafted features such as Scale Invariant Feature Transform (SIFT) and Histogram of Oriented Gradients (HOG). In the last ten years convolution networks have been proposed to extract features from faces~\cite{Zhang}. 
%The core component of these networks is convolution

% \begin{equation}
% (f \star h)(i,j) = \sum_{k=-T}^T \sum_{m=-T}^T f(k,m) \cdot h(i - k, j - m),
% \end{equation}
% where $f(x)$ is a kernel function whose parameters are learnt from the data of the task in hand. 

For our purposes, we utilise a deep residual network (ResNet) of 50 layers~\cite{he2016deep}. As input to the network we use the raw pixel intensities from the cropped faces of a subject's video. 
% Deep residual networks adopt residual learning by stacking building blocks of the form: $\bs y_k = \mathcal{F}( \bs x_k,\{W_k\}) + \bs h(x_k)$,
% where $\bs x$ and $\bs y$ are the input and output of the layer $k$, $\mathcal{F}( \bs x_k,\{W_k\})$ is the residual function to be learnt, and $h(\bs x_k)$ can be either an identity mapping or a linear projection to match the dimensions of the function $\mathcal{F}$ and the input $\bs x$.
The architecture of the network starts with a $7 \times 7$ convolutional layer with 64 feature maps, followed by a max pooling layer of size $3 \times 3$. Then, four bottleneck architectures are utilised to extract higher-level abstractions where after these architectures a shortcut connection is added. Each of these architectures is comprised of three convolutional layers of sizes $1 \times 1$, $3 \times 3$, and $1 \times 1$, for each residual function. Table~\ref{bottleneck_architectures} shows the replication and the sizes of the feature maps for each bottleneck architecture. The last layer of the network is an average pooling.

\begin{table}[ht]
\centering
\caption{The replication of each bottleneck architecture of the ResNet-50 along with the size of the features maps of the convolutions.}
\begin{tabular}{l|l|l}\toprule\midrule
 Bottleneck layer & Replication & \parbox[t]{3cm}{\# Feature maps\\ } \\\midrule
 1st & 3 & 64, 64, 256 \\
 2nd & 4 & 128, 128, 512 \\
 3rd & 6 & 256, 256, 1024 \\ 
 4th & 3 & 512, 512, 2048 \\ 
\end{tabular}

\label{bottleneck_architectures}
\end{table}

% \subsection{Contextual Extractor}
% \label{sec:rnn}
\textit{Contextual Extractor}: The spatial-modelling of the signal removes background noise and enhances specific parts of the signal in our task, but it does not model the temporal structure of sequential data such as the videos in our dataset. To model such structure, we feed the extracted facial features to a 2-layer LSTM network.

\section{Model Pretraining}

In this section, we describe the dataset provided by RealEyes (Sec.~\ref{sec:dataset}). In addition, we present the training methodology used to tackle the unbalanced data (Sec.~\ref{train_process}) along with the loss function to train the ResNet-50 model (Sec.~\ref{loss_func}). 

\subsection{RealEyes Dataset}
\label{sec:dataset}

The model was trained on a dataset provided by the RealEyes company that specialises on recognising human emotions. The dataset is comprised of videos collected by individuals from a personal computer camera recording while they were watching an advertisement. Each video contains one participant. The aim was to collect spontaneous facial behaviour in arbitrary recording conditions. We should point out that no audio information is available in the videos due to privacy issues. The dataset is comprised of $4,973$ videos, with a total length of more than 31 hours. The total number of subjects is $2,616$, with $1,099$ being males and $1,517$ females. Table~\ref{tab:dataset} shows a summary of the statistics of the dataset.

Each frame of the dataset is annotated by seven expert annotators~\footnote{All annotators have advanced understanding on face analysis problems, and in particular of facial expressions.} and for eight emotions, namely, neutral, happy, surprise, disgust, contempt, confusion, empathy, and other. In particular, all annotators were instructed both orally and through a document on the process to follow for the annotation. This document included a list of some well identified emotional cues for all emotions, providing a common basis for the annotation task. On top of that, the experts used their own appraisal of the subject’s emotional state for creating the annotations. 

The annotation process provided the freedom to the annotators to watch the videos, pause, re-wind, and then mark the start of an emotional state. The final label of the frame is given by a winner-takes-all approach, meaning the maximum number of votes for each emotion. Table~\ref{tab:emo_stat} shows the number of frames per emotion.

\begin{table}[ht]
\centering
\begin{tabular}{l|r}\toprule\midrule
 Attribute    & Value \\\midrule
\# Videos     & 4,973 \\
\# Frames     & 1,059,505 \\
Frames per video $\mu(\sigma)$ & $380 (308.39)$ \\
Image Resolution & $640 \times 480$ \\ \hline
\# Subjects   & 2,616 \\
Females/Males   & 2,983/1,990 \\
Age variation & [18-69] \\ \hline
\# Annotators & 7 \\
\# Emotions & 8 \\
\end{tabular}
\caption{The dataset statistics provided by RealEyes.}
\label{tab:dataset}
\end{table}

\begin{table}[ht]
\centering
\begin{tabular}{l|r}\toprule\midrule
 Emotion    &  \# Frames \\\midrule
Happy & 60,179\\
Surprise & 15,786\\
Confusion & 56,683\\
Disgust & 10,065 \\
Contempt & 8,581\\
Empathy & 10,173\\
Neutral & 840,611 \\
Other & 57,425 \\\midrule
Total Count &   1,059,505 \\
\end{tabular}
\caption{The number of frames for each emotion, provided by RealEyes.}
\label{tab:emo_stat}
\end{table}

\subsection{Training Process}
\label{train_process}

The input to our network are the raw pixel intensities from the face extracted from the frames of the videos using the Multi-Domain Convolutional Neural Network Tracker (MDNet)~\cite{nam2016learning} tracking algorithm. The extracted frames of the face are then resized to resolution $150 \times 150$.

A major problem that occurs when trying to train our network with this dataset, is the highly unbalanced labels. Most of the labels are neutral ({$\approx 78$ \%}), and this poses the threat of the network to classify all of the frames as neutral. To deal with this, we use a boosting method by re-sampling the 
frames that contain labels with the lowest portion in the dataset, namely, happy, surprise, disgust, confusion, empathy, and contempt. In particular, we re-sampled window video frames which contain the least frequent emotions and at the same time do not contain frequent emotions such as neutral. The window length was chosen to be $150$ frames same as in our training framework. 

An important issue that occurs is that emotions are temporally expressed. That is, in a sequence of frames. Hence, our re-sampling strategy is based on this fact and we sampled sequences of frames instead of just individual frames. After this process, we randomly split the dataset based on the participants to training and validation set ($80\%$ training, $20\%$ validation). Table~\ref{tab:emo_stat_oversample} shows the number of frames per emotion after the re-sampling.

\begin{table}[ht]
\centering
\begin{tabular}{l|r|r}\toprule\midrule
 Emotion    &  \# Training Frames &  \# Validation Frames \\\midrule
Happy & 298,222 & 9,494\\
Surprise  & 237,041 & 1,518 \\
Confusion & 265,495 & 9,377 \\
Disgust & 32,044 & 346\\
Contempt & 18,537 & 569\\
Empathy & 21,444 & 745\\
Other & 52,375 & 1,703\\
Neutral & 877,410 & 18,895\\\midrule
Total Count & 1,505,443 & 43,492 \\
\end{tabular}
\caption{The number of frames of each emotion after oversampling, provided by RealEyes.}
\label{tab:emo_stat_oversample}
\end{table}

\subsection{Experimental Setup}

We utilised the Adam optimisation algorithm to train the model  with a fixed learning rate of $4*10^{-5}$. We used a mini-batch of size $2$ with a sequence length of $150$ frames. For our purposes we used data augmentation methods. In particular, the data were augmented by resizing the image to size $170 \times 170$ and randomly cropping it to equal its original size. With this method, the model becomes scale invariant. In addition, colour augmentation is used by introducing random brightness and saturation to the image. Finally,  to speed up the convergence of the optimisation algorithm we initialise the weights of our feature extractor using the pre-trained model on ImageNet~\cite{deng2009imagenet}. 
\\
\subsection{Loss Function}
\label{loss_func}

We should note that each frame in our dataset contains multiple labels, and as such our loss function is formulated as follows:

\begin{equation}
    \mathcal{L} = - \bs z * \log(\sigma( \bs x)) + (\bs 1 - \bs z) * \log(1 - \sigma(\bs x)),
\end{equation}

where $\bs z$ are the labels and $\bs x$ the predictions. We should point out that we use a sigmoid function ($\sigma$) because we want to predict multiple emotions (labels) for each frame. Finally, we weight each sample based on the reciprocal of the number frames of each emotion. Hence, the emotion with the lowest number of samples (Contempt) has the highest weight, while the emotion with the highest number of frames (Neutral) the lowest weight.

\section{Experiments}
\label{sec:experiments}

% We test the model trained on the RealEyes dataset on a common emotion recognition database, namely, the REmote COLlaborative and Affective (RECOLA) (Section~\ref{recola}). As loss function we use the Concordance Correlation Coefficient ($\rho_c$) (Section~\ref{ccc}). By initialising the model with our pre-trained one, we compare with models that their weights are initialised randomly, and using the pre-trained model on ImageNet. Our results (Section~\ref{results}) show the superiority of our pre-trained model.

\subsection{Dataset}
\label{recola}

To test our pre-trained model, we utilise the REmote COLlaborative and Affective (RECOLA) database introduced by Ringeval et al.~\cite{ringeval2013introducing}. A part of the database was used in the Audio/Visual Emotion Challenge and Workshop (AVEC) 2016-2018 challenges~\cite{valstar2016avec}. For our purposes, we utilise the full database, which contains 46 different recordings divided into three different parts (train, devel, and test) while balancing the gender, age, and mother tongue. In total, four modalities are contained in the database, i.\,e., audio, video, electrocardiogram (ECG), and electro-dermal activity (EDA). The original labels of the RECOLA are re-sampled at a constant frame rate of 40\,ms. The data is then averaged over all raters by considering the inter-evaluator agreement, to provide a ‘gold standard’~\cite{han2017reconstruction}. Each record is 5\,min (300\,s) audio data with a sampling rate of $16$\,kHz.

% Table~\ref{tab:recola_dataset} contains more details for each portion of the dataset.

% \begin{table}[htbp]
% \centering
% \caption{The partitioning of the RECOLA dataset.}
% \begin{tabular}{|l|l|l|l|}
% \hline
%                 & \textbf{Train} & \textbf{Valid} & \textbf{Test} \\ \hline
% female/male     & 10/6           & 9/6            & 8/7           \\ \hline
% French          & 11             & 11             & 11            \\ \hline
% Italian         & 3              & 2              & 3             \\ \hline
% German          & 2              & 1              & 1             \\ \hline
% Portuguese      & 0              & 1              & 0             \\ \hline
% age $\mu(\sigma)$ & 22.3(3.4)      & 21.6(2.1)      & 21.2(2.0)     \\ \hline
% \end{tabular}
% \label{tab:recola_dataset}
% \end{table} 

\subsection{Experimental Setup}

The optimisation method we used to train our model, throughout all experiments, is the Adam optimisation method with a fixed learning rate of $4*10^{-5}$. We used a mini-batch of size $2$ with a sequence length of $150$ frames ($12$\,s) - total $300$ frames per batch - in both training and testing. Our loss function is based on the concordance correlation coefficient ($\rho_c$)~\cite{ringeval2017avec}, and the best model is selected based on the highest $\rho_c$ on the validation set. The number of epochs was set to $300$. We stopped the training process if no improvement on the validation score was observed after 20 epochs. Finally, a chain of post processing methods is applied: (a) median filtering (the size of the window was between 0.04\,s and 20\,s)~\cite{valstar2016avec}, (b) centring (by finding the ground truth's and the prediction's bias)~\cite{kachele2015ensemble}, (c) scaling (with the scaling factor the ratio between the standard deviation of the ground truth and the prediction)~\cite{kachele2015ensemble}, and (d) time-shifting (forward in time with values between 0.04\,s and 10\,s). Any of these methods is kept when we observe a better $\rho_c$ on the validation set, and then applied to the test partition with the same configuration.

% \subsection{Loss Function}
% \label{ccc}

% To train the models our loss function is based on the concordance correlation coefficient ($\rho_c$), which has been shown to provide better results~\cite{trigeorgis2016adieu} than the traditional Mean Squared Error (MSE). Similar to previous studies, we calculated $\rho_c$ on pooled dataset (ignoring that we have multiple people and multiple observations per person).

% Mathematically, the $\rho_c$-based loss function ($\mathcal{L}_c$) is computed as follows:

% \begin{align}
% \mathcal{L}_c = &1 - \rho_c = 1 - \frac{2\sigma_{xy}^2}{\sigma_x^2+\sigma_y^2+(\mu_x - \mu_y)^2} \\
%             = &1 - 2\sigma_{xy}^2\psi^{-1},
% \end{align}
% % 
% %  

% where $\psi = \sigma_x^2+\sigma_y^2+(\mu_x - \mu_y)^2$ and $\mu_x = E(x)$,$\mu_y = E(y)$,$\sigma^2_x=var(x)$,$\sigma^2_y=var(y)$ and $\sigma_{xy}^2 = cov(x,y)$. 
% The gradient of the loss to be propagated from the last layer with respect to the weights is

% \begin{align}
% \frac{\partial\mathcal{L}_c}{\partial x} \propto 2 \frac{\sigma_{xy}^2(x-\mu_y)}{\psi^2} + \frac{\mu_y - y}{\psi}.
% \end{align}

% We use $\rho_c$ as our evaluation metric. As our models were trained to output both arousal and valence our overall loss is defined as:

% \begin{equation}
%     \mathcal{L}_c = \frac{\mathcal{L}_a + \mathcal{L}_v}{2},
% \end{equation}

% \noindent
% where $\mathcal{L}_a$ and $\mathcal{L}_v$ are the loss values for the arousal and the valence, respectively.

\subsection{Results}
\label{results}

\begin{table}[tbp]
\centering

\begin{tabular}{c|c|c} \toprule\midrule
Initialisation Strategy & Arousal      &  Valence \\ \midrule
  Random              & .319~(.342)   & .481~(.548)            \\ 
Pre-trained (ImageNet) & .324~(.330)   & .492~(.554)            \\ 
AffWildNet~\cite{kollias2018deep} & .273~(--)   & .526~(--)            \\ \midrule
Proposed model   & \textbf{.341}~(.376) & \textbf{.538}~(.596) \\ 
\end{tabular}
\caption{Performance on the RECOLA dataset (w.\,r.\,t.\ $\rho_c$) between the proposed model and the rest. In parenthesis are the performance obtained on the validation set. A dash is inserted if the results could not be obtained.}
\label{tab:compare_models}
\end{table}

We fine-tune the model trained on the RealEyes dataset on the RECOLA database. For our training purposes we do not use any kind of data augmentation, such as colour distortion or random cropping. We compare the model with models initialised using two approaches: (i) from scratch, i.\,e. randomly using Xavier initialisatin, (ii) using the weights of the model pre-trained on the ImageNet dataset, and (iii) the current state-of-the-art AffWildNet~\cite{kollias2018deep}. The same training framework was used to all models, i.\,e., a feature- and a contextual-extractor. Table~\ref{tab:compare_models} shows the results. 

As can be observed, our model provides the best results compared to the other methods in both the arousal and valence dimensions, which indicates the superiority of the pre-trained model. We should also point out that our pre-trained model has faster convergence (i.\,e., the time to reach its best performance) of the optimisation than the other two models. In particular, our model convergence's is more than 2 times faster than the pre-trained one from ImageNet, and almost 3 times faster than the random initialisation. Finally, we should note that all the models perform better in the valence dimension rather than the arousal one, which has also been shown in other emotion recognition studies that utilise the face for dimensional emotion recognition~\cite{tzirakis2018end2you}.

\section{Conclusions and Future Work}
\label{sec:concls}

In this paper, we proposed a pre-trained facial feature extractor, namely, a ResNet-50 model, which was trained on a large in-the-wild dataset provided by the RealEyes company with eight emotions. Before training the model, we used a boosting technique by resampling the dataset due to the highly unbalanced labels. 
Although the model was trained for categorical emotion recognition, we showed that fine-tuning it on the RECOLA database provides the best results on both arousal and valence dimensions when comparing it with models whose weights are initialised (i) randomly, (ii) using the pre-trained model on ImageNet, and (iii) using the current state-of-the-art AffWildNet~\cite{kollias2018deep}. Our model surpasses the performance of the other models in terms of the concordance correlation coefficient, thus, making it an excellent tool for recognising facial affect. 

For future work, we intend to perform more experiments on other dimensional emotion recognition datasets such as on the cross-cultural Sentiment Analysis in the Wild (SEWA) database. In addition, we will include categorical emotion recognition datasets 

% such as the AFEW dataset~\cite{kossaifi2017afew} that comprises of seven basic emotions. Finally, it would be interesting to experiment on tasks different from emotion recognition such as face identification.

% \section*{Acknowledgments}

\vfill\pagebreak

% \section{REFERENCES}
% \label{sec:refs}

\bibliographystyle{IEEEbib}
\bibliography{refs}

\end{document}